\documentclass{article}
\usepackage{url}
\usepackage{latexsym}
\usepackage{natbib}
\renewcommand{\cite}[1]{\citep{#1}}

\usepackage{microtype}
\usepackage{graphicx}
\usepackage{subfigure}
\usepackage{booktabs}

\usepackage{algorithm}
\usepackage{algorithmic}
\usepackage{amsmath}
\usepackage{amssymb}
\usepackage{bm}
\usepackage{upgreek}
\usepackage{amsthm}
\usepackage{caption}
\usepackage{wrapfig}
\usepackage{standalone}
\usepackage{tikz}
\usepackage{pgfplots}
\pgfplotsset{compat=1.16}
\usetikzlibrary{positioning,shapes}

\usepackage{glossaries}
\glsdisablehyper
\newacronym{nlp}{NLP}{Natural Language Processing}
\newacronym{rnn}{RNN}{Recurrent Neural Network}
\newacronym{cnn}{CNN}{Convolutional Neural Network}
\newacronym{tnn}{TNN}{Tree Neural Network}

\newcommand{\bigO}{\mathcal{O}}

\usepackage{hyperref}

\title{An Iterative Contextualization Algorithm\\with Second-Order Attention}

\author{Diego Maupomé, 
        Marie-Jean Meurs \\
        Universit\'e du Qu\'ebec \`a Montr\'eal\\
        {\small{\tt maupome.diego@courrier.uqam.ca}}\\
        {\small{\tt meurs.marie-jean@uqam.ca}} \\
  }

\date{~}

\begin{document}

\maketitle

\begin{abstract}
Combining the representations of the words that make up a sentence into a cohesive whole is difficult, since it needs to account for the order of words, and to establish how the words present relate to each other.
The solution we propose consists in iteratively adjusting the context. 
Our algorithm starts with a presumably erroneous value of the context, and adjusts this value with respect to the tokens at hand. 
In order to achieve this, representations of words are built combining their symbolic embedding with a positional encoding into single vectors. 
The algorithm then iteratively weighs and aggregates these vectors using our novel second-order attention mechanism. Our models report strong results in several well-known text classification tasks.
\end{abstract}

\section{Introduction}
\label{sec:intro}

Until recently, encoding natural language utterances was based on some aggregation of the parts of the utterance where the representation of such parts was constant, independent of the utterance.
Even non-symbolic representations, such as word embeddings, were not dependent on the utterance that they might comprise.
That is, the \emph{context} of each word only appeared \emph{a posteriori}.
This made the issue of polysemy difficult to address.

However, in the case of some neural encoders, such as \glspl{rnn}~\cite{jordan1986attractor}, one might argue that some notion of contextualization does exist.
In \glspl{rnn}, as the sequence is consumed, more information about the sequence becomes available to the hidden state updates.
Obviously, this is highly dependent on the parsing order, and the significance of certain tokens has to be presupposed to some extent. 
For example, when reading a phrase one word at a time, the significance of the first word is at first supported only by prior beliefs.
Without actually reading the entire phrase, one can only suppose what the role of the first word will be.
The second and third word will be better informed, but still somewhat dependent on prior beliefs.
Only the last word of a sentence will be fully "aware" of its role, memory problems notwithstanding.
Bidirectional \glspl{rnn}~\cite{schuster1997bidirectional} mitigate these issues by offering two parsing orders to sequences.
However, the contexts compiled by the hidden states of the forwards and backward \glspl{rnn} are not aware of one another, they are not \emph{revisited}.

By contrast, Transformer encoders~\cite{vaswani2017attention} readjust the representation of words iteratively.
The words in a sequence are compared against each other by the parametric Self-Attention mechanism~\cite{cheng2016long}.
From this process, a new representation for each word is produced.
This is repeated a set number of times.
In doing so, the word representations produced by this encoder are put into the context of the whole, ostensibly addressing issues of ambiguity or polysemy.
As such, Transformers have achieved much success in various \gls{nlp} tasks~~\cite{devlin2018bert,cer2018universal,clark2020electra}.
Nonetheless, the Self-Attention mechanism on which Transformers rely has two chief disadvantages.
Firstly, because all word pairs are evaluated, the complexity is quadratic with respect to the length of the utterance.
Secondly, the weighting provided by the Self-Attention mechanism is based on bilinear forms, mapping each pair of word vectors to a single scalar.
As such, Transformers require multiple sets of Self-Attention parameters called \emph{heads}, so that separate heads might focus on different features of the word vectors.
To address these issues, we propose a novel architecture - the \textit{Contextualizer} - based iteratively adjusting a context vector using a second-order attention mechanism.

This paper is organized as follows.
Section \ref{sec:context} introduces the proposed approach.
Section \ref{sec:expes} describes experiments conducted in a few well-known document classification tasks and the results obtained. 
Finally, Section \ref{sec:conclusion} concludes this paper.

\section{Contextualizer}
\label{sec:context}

Contextualization is fundamentally difficult because it is a circular problem.
One cannot recompose a whole by putting its tokens in the context of said whole without already knowing what the whole is.
One potential solution to this is to iteratively adjust the context.
That is, to begin with a presumably erroneous value of the context representation and adjust this value with respect to the tokens at hand.

In order to achieve this, representations of words are built combining their symbolic embedding with a positional encoding into single vectors. 
The algorithm then iteratively weighs and aggregates these vectors using our novel second-order attention mechanism.
In broad terms, the algorithm we propose is presented in Algorithm~\ref{alg:cntxt}.

\begin{algorithm}
\caption{Contextualizer}
\label{alg:cntxt}
    \textbf{Input} $C$: a set of tokens to be contextualized,
    \\$c_d$: a default context vector\\
    \textbf{Output} $c$: a context vector
\begin{algorithmic}
    \STATE $c \gets c_d$
    \LOOP
        \STATE $C' \gets \varnothing$
        \FORALL{$t \in C$}
            \STATE $C' \gets C' \cup \{contextualize(c, t)\}$
        \ENDFOR
        \STATE $c \gets aggregate(C')$
    \ENDLOOP
\end{algorithmic}
\end{algorithm}
\vspace{.4cm}

We have established that our approach requires a contextualizing and an aggregating function.
We use a variant of the attention mechanism to implement contextualization and simple addition for aggregation.
We begin by describing the representation of tokens as follows.  

Let $w_1,\ldots,w_n$ be a sequence of tokens forming a document of length $n$.
An encoding function, $e$, maps each token and its position to a single real vector, $\bm{x}_i\in \mathbb{R}^m$,

\begin{equation}\label{eq:inp}
    e:(w_i, i)\mapsto \bm{x}_i.
\end{equation}
This function can be learned or manually specified.

Suppose there are $K$ contextualization and aggregation steps, indexed by $k=1,\ldots,K$.
Each of these steps will produce a new context vector, $\bm{c}^k$.
As described by Algorithm~\ref{alg:cntxt}, this context vector will contextualize the tokens, which will then be aggregated into a new context.
An attention mechanism provides the contextualizing function called at every iteration.
Using any of the various attention mechanisms in the literature, contextualizing each token would amount to producing a scalar weight, $\alpha_i$ for each token depending on its content and that of the attender (the context vector in our case).
The contextualization of token $\bm{x}_i$ at step $k$ with respect to the previous context, $\bm{c}^{(k-1)}$, would then be

\begin{equation*}
\label{eq:scalarweight}
    (\bm{c}^{(k-1)}, \bm{x}_i) \mapsto \alpha^{(k)}_i\bm{x}_i.
\end{equation*}

However, we would prefer to have the weight of each token be a vector, $\boldsymbol{\alpha}_i$, rather than a scalar, $\alpha_i$.
This would let each component of the token representation have a separate salience with regard to the current context.
This is particularly important when using distributed token representations, where each component might carry a different semantic sense.
Using a weighting vector, the previous mapping would become:

\begin{equation}
    (\bm{c}^{(k-1)}, \bm{x}_i) \mapsto  \boldsymbol{\alpha}^{(k)}_i*\bm{x}_i
\end{equation}

where $*$ denotes the Hadamard product.
This is an appealing alternative to the use of separate Self-Attention \emph{heads}, as prevalent in Transformer models.
Indeed, the use of scalar attention weights requires that each component in the operands interact only with its homologue, collapsing all information to a single number.
One must therefore compute several of these interaction with different parameters, heads, so that each of these heads may focus on different features.
A second-order attention mechanism eliminates the need for several heads, as each feature can interact with each feature by a different parameter. 

Nonetheless, this second-order attention weighting requires parametrization by a tensor of degree 3.
This would take the parameter count of the model to $O(m^3)$, as both the input and the context vectors are of dimension $m$.
For token representations of even modest size, this would result in a computationally intensive model that would ostensibly be prone to over-fitting.
Instead, we can compose a degree 3 tensor of arbitrary rank $u$ by adding together $u$ degree 3 tensors of rank 1.
These can be expressed by a three-way outer product of vectors~\cite{rabanser2017introduction,sutskever2011generating,maupome-meurs-2020-language}.

\begin{figure}
    \caption{Computation of the weight vectors factored into three parameter matrices}
    \label{fig:weight}
    \vskip 0.15in
    \centering
    \includegraphics[width=.64\textwidth]{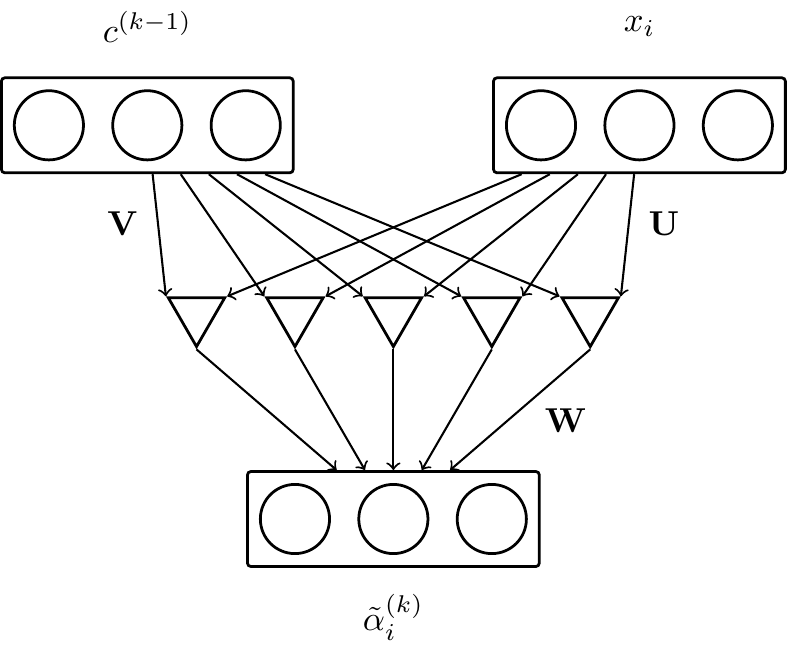}
\end{figure}

The candidate weighting of each token can therefore be computed as follows:

\begin{equation*}\label{eq:ztc}
    \tilde{\boldsymbol{\alpha}}^{(k)}_i = \mathbf{W}^{(k)} (\mathbf{U}^{(k)}\bm{x}_i * \mathbf{V}^{(k)} \bm{c}^{(k-1)}),
\end{equation*}

where $\mathbf{U}^{(k)},\mathbf{V}^{(k)}\in \mathbb{R}^{u\times n}$ and $\mathbf{W}^{(k)}\in \mathbb{R}^{n\times u}$ are the matrices of parameters for the $k$th contextualization step.

This computation is illustrated in Figure~\ref{fig:weight}.

These candidate weights can then be normalized across tokens to provide the true weights for each token:

\begin{equation*}\label{eq:softmax}
    \alpha_{ij}^{(k)} = \frac{\mathrm{exp}(\tilde{\alpha}^{(k)}_{ij})}{\sum_{l=1}^n \mathrm{exp}(\tilde{\alpha}^{(k)}_{lj})}
\end{equation*}

The context is then obtained by adding the weighted tokens together:

\begin{equation*}\label{eq:agg}
    \bm{c}^{(k)} = \sum_{i=1}^n \boldsymbol{\alpha}^{(k)}_i*\bm{x}_i
\end{equation*}

The default context used at the first step of contextualization, $\bm{c}^{(0)}$, can be set to a constant, for example. 
Furthermore, we call \emph{recurrent contextualizers} models where the same matrices are used for all iterations: 
\begin{equation*}\label{eq:mat}
(\mathbf{W}^{(k)}, \mathbf{U}^{(k)}, \mathbf{V}^{(k)}) = (\mathbf{W}, \mathbf{U}, \mathbf{V}) \ \text{for} \ k=1,\ldots,K.
\end{equation*}
\vspace{.1cm}
where $K$ is the number of contextualization steps.
\vspace{.2cm}

An alternate approach to recontextualization is to iteratively adjust the tokens with respect to one another.
This algorithm is presented in Algorithm~\ref{alg:compcntxt}.
It can be understood as the Encoder portion of the popular Transformer architecture~\cite{vaswani2017attention}, where several layers of self-attention are applied for each token in a sentence to arrive at a contextualized representation.
It has the advantage of decentralizing the context.
This allows for the context to be of variable size in a natural way.\\

\begin{algorithm}
\caption{Token-wise Contextualizer}
\begin{algorithmic}
    \LOOP
        \FORALL {$t \in C$}
            \STATE $C' \gets \varnothing$
            \FORALL{$t \in C$}
                \STATE $C' \gets C' \cup \{contextualize(t, t')\}$
            \ENDFOR
            \STATE $t \gets aggregate(C')$
        \ENDFOR
    \ENDLOOP
\end{algorithmic}
\label{alg:compcntxt}
\end{algorithm}
\vspace{.2cm}

The Contextualizer, on the other hand, has a fixed-size context and must additionally provide a default context.
In return, the number of comparisons in one iteration of the Contextualizer algorithm grows linearly with respect to the number of tokens, as opposed to the square of it for the Transformer.
Table \ref{tab:complexities} presents the complexities of these two algorithms as well as recurrent and convolutional layers.
In addition, the Transformer has the drawback of losing sight of the original representation of the tokens, whereas the Contextualizer does not.
In the next section, we present a few text classification experiments in order to analyze the proposed Contextualizer approach.

\begin{table}
    \centering
    \begin{tabular}{l|c}
        Layer & Complexity\\
        \hline
        Recurrent & $\bigO(nm^2)$\\
        Convolutional & $\bigO(knm^2)$\\
        Transformer Encoder & $\bigO(hn^2m)$\\
        Contextualizer & $\bigO(num)$\\
    \end{tabular}
    \caption{Complexities of common layers used in NLP, $k$ designates the kernel size, $h$, the number of attention heads, $n,m$ and $u$, designate the length of the sequence, the dimension of the word representations and the multiplicative dimension, respectively. For Transformers and Contextualizers, the complexity is for a single contextualization step.}
    \label{tab:complexities}
\end{table}

\section{Experiments and Results}
\label{sec:expes}

We evaluate the Contextualizer approach on several text classification tasks.
We begin with exploratory experiments and analysis on a well-known sentiment analysis dataset:asdf.
We then present further results in other text classification benchmarks.

For all of these tasks, we opt for word-level tokenization.
Words appearing fewer than 3 times are removed.
As described by Equation~(\ref{eq:inp}), both the position and the value of a word are to be combined into a single real vector.
The position of the word is encoded and integrated to the word embedding as described by \citet{vaswani2017attention}.
In order to isolate the learning capabilities of the Contextualizer, we do not use pre-trained word embeddings or learn them during the task.
Instead, each word is associated with a random vector, $\bm{w} \sim \mathcal{U}(-\mathbf{1}, \mathbf{1})$.
Neither the value or the position vector encoding of words is trained.
The Contextualizer must therefore learn to work with the word representations it is given.

The results reported issue from 5-fold cross-validation.
Models are trained over 10 epochs in batches of 64 documents using the Adam optimizer~\cite{adam}.
Accuracy is used to measure the models performance, a natural choice given that all tasks are binary classification tasks with equal number of observations per class.
We retain the models with the best performance on a development set constituting 10\% of each training set fold.
All data subsets are constructed randomly and respecting the balance between labels. 
Naturally, the results reported represent the test accuracy averaged across the folds.

\subsection{Exploratory experiments}

We begin with experiments on the well-known Rotten Tomatoes  dataset (MR)~\cite{pang2005seeing}.
The dataset consists of 11k sentences from film reviews classified as either positive or negative in equal proportions.
Removing words having fewer than 3 occurrences results in a vocabulary of about 5700 words.
The documents are fairly short, with 95\% of them being 45 words long or shorter.

Our first experiments compare how the number of contextualization steps, $K$, affects performance in a Contextualizer.
We train models with the aforementioned configurations with varying depth of contextualization. 
As such, $K$ takes on the values $1, 5, 10$ and $20$.
We train both recurrent and regular models with these settings.

Because the word representations are not trained, the actual number of parameters for recurrent Contextualizers is fairly small, counting only the three attention matrices and the parameters of the affine transformation mapping the final context to a binary decision for the document.

We set $v$, the size of the word embeddings, to 500, and $p$, the size of the position vector to 20.
The rank of the 
This yields a total parameter count for the models to 157k.
In the case of recurrent models, the choice of $K$ does not affect the parameter counts because the same matrices are used for every step.
For non-recurrent models, the parameter count goes up to 3M.

Results are presented in Table \ref{tab:depth}.
The depth of contextualization appears to have little effect for $K \geq 5$ for recurrent and non-recurrent models alike.
Moreover recurrence seems to have no bearing on performance, suggesting that the \hbox{resulting} models are robust to over-fitting.

\begin{table}
\caption{Test accuracy (\%) on the MR task for varying number of contextualization steps ($K$)}
\label{tab:depth}
\vskip 0.15in
    \centering
    \begin{tabular}{c|c c}
        $K$ &recurrent &regular \\
        \hline
        1 & 57.9 &57.8\\
        5 & 72.4 &70.7\\
        10 & 72.8&72.1\\
        20 & 72.3&71.4\\
    \end{tabular}
\end{table}

\begin{table}
    \caption{Test accuracy (\%) on the MR task for different default context strategies}
    \label{tab:default}
    \vskip 0.15in
    \centering
    \begin{tabular}{c|cc}
        &\multicolumn{2}{c}{$K$}\\
        $\bm{c}^{(0)}$ & 1 & 5\\
        \hline
        $\mathbf{1}$&73.1 & 71.2\\
        $\bm{c}_d$&73.5 & 72.2\\ %
        $\mathcal{U}(-\mathbf{1}, \mathbf{1})$ &57.9 & 72.4\\ %
    \end{tabular}
\end{table}

We then proceed with experiments measuring the effect on performance of the nature of the default context.
All models share the same configurations except the default context, which is set to be either a constant, $\bm{c}^{(0)}=\mathbf{1}$, a vector of learned parameters, $\bm{c}^{(0)}=\bm{c}_d$, or a random vector redrawn for every document, $\bm{c}^{(0)}\sim \mathcal{U}(-\mathbf{1}, \mathbf{1})$.
We hypothesize that using a random default context will make the network more robust by reducing dependence on prior beliefs and therefore mitigating overfitting.
For the same reasons, one could expect a learned default context to be more likely to overfit than a constant one.
We run these experiments on recurrent models with $K$ set to 1 and 5.

Table \ref{tab:default} summarizes the results.
As one might expect, a random starting context vector hurts performance when contextualization is performed but once.
The models are quick to adjust, as all choices of default context seem to arrive at very similar final accuracies.

\subsection{Further results}

We continue with experiments in binary document classification on other well-known datasets. 
The Customer Reviews dataset (CR), introduced by \citet{hu2004mining}, comprises 3775 reviews roughly equally divided between positive and negative sentiment.
The Subjectivity dataset (SUBJ)~\cite{pang2004sentimental} groups 10k sentences as subjective or objective.
Finally, the Multi-Perspective Question Answering classification dataset (MPQA) deals again in sentiments polarity, with 10,606k phrases.

Once words with fewer than 3 occurrences are removed, the largest vocabulary is that of SUBJ, with about 6.3k words, followed closely by MR, with 5.7k words.
The two other datasets have much smaller vocabularies, with 1.7k and 1.5k words for CR and MPQA, respectively.
The MPQA dataset is the most unbalanced in terms of labels found, having a ratio of 7 to 3 negative to positive sentences.
Still, both cross-validation as well as the sampling of developments subsets are performed in a stratified manner, preserving the proportions between classes.

We also test the gains in performance offered by jointly learning the word embeddings.
However, in order to offset the increase in parameter count learning these embeddings entails, we reduce the size of the word vectors by half, to 250.
This results in models of size with about half a million parameters for CR and MPQA and a million and a half for SUBJ and MR.
The default context is again random, with 5 contextualization steps adjusting it.

Test accuracy on several benchmark text classification tasks of our Contextualizer models compared to the Universal Sentence Encoder architectures (USE)~\cite{cer2018universal}, Transformer-based (T) and Deep Averaging Network-based (D) is shown in Table \ref{tab:further}. 
The results of these experiments demonstrate that the Contextualizer architecture can perform competitively, even with small models and relatively small datasets.
More importantly, the gains in performance of learning word representation are relatively small, suggesting that the heavy lifting is done by the higher-order attention function.

\begin{table}
\caption{Test accuracy (\%) on several benchmark text classification tasks of our Contextualizer models compared to the Universal Sentence Encoder architectures (USE), Transformer-based (T) and Deep Averaging Network-based (D)}
    \label{tab:further}
    \vskip 0.2in
    \centering
    \begin{tabular}{c|cccc}
         & MR & CR & SUBJ & MPQA\\
         \hline
        Contextualizer & 76.6 & 79.0 & 91.2 & 85.3\\
        USE (T) & 81.4 & 87.4 & 93.9 & 87.0\\
        USE (D) & 74.5 & 81.0 & 92.7 & 85.4
        
    \end{tabular}
\end{table}

\section{Conclusion}
\label{sec:conclusion}

We have proposed an algorithm for constructing sentence representations based on the notion of iteratively adjusting a central context vector.
This algorithm is closely related to the encoder part of the Transformer algorithm.
One key difference is the use of the proposed second-order attention mechanism, replacing multiple attention heads.
Our results suggest the approach is robust to different choices of the number of contextualization steps and default contexts.
Furthermore, the Contextualizer can achieve competitive results in benchmark document classification tasks even with low parameter counts.

Transformer models have been the driving force behind the expansive use and development of large models such as BERT~\cite{devlin2018bert} and GPT-2~\cite{radford2019language}, which are extensively trained by adapted language-modeling tasks.
The reduced complexity of the Contextualizer model would be of use both in terms of pre-training and in terms of wielding these large models in downstream tasks.
Further work will be conducted in this direction as well as in formally characterizing the conditions that stabilize the context vector as the number of contextualization steps increases.\\

\section*{Reproducibility}
The Contextualizer source code is available under the GNU GPL v3 license to ensure reproducibility.
It can be found in the following repository:\\ {\footnotesize{\url{https://gitlab.ikb.info.uqam.ca/ikb-lab/nlp/contextualizer/Contextualizer_arXiv2021}}}

\section*{Acknowledgements}
This research was enabled in part by support provided by Calcul Québec ({\footnotesize{\url{https://www.calculquebec.ca}}}) and Compute Canada ({\footnotesize{\url{https://www.computecanada.ca}}}). 
We acknowledge the support of the Natural Sciences and Engineering Research Council of Canada (NSERC) [MJ Meurs, NSERC Grant number 06487-2017], and the support of the Government of Canada’s New Frontiers in Research Fund (NFRF), [MJ Meurs, NFRFE-2018-00484].

\bibliographystyle{apalike}
\bibliography{maupome_meurs_contextualizer_2021}

\end{document}